%% file: nips2016.tex
\pdfoutput=1
\documentclass{article}

\usepackage[final]{nips_2016}

\usepackage{natbib}
\usepackage[utf8]{inputenc} 
\usepackage[T1]{fontenc}    
\usepackage{hyperref}       
\usepackage{url}            
\usepackage{booktabs}       
\usepackage{amsfonts}       
\usepackage{nicefrac}       
\usepackage{microtype}      
\usepackage{graphicx}
\usepackage{multirow}
\usepackage{amsmath, amssymb}

\title{A Binary Convolutional Encoder-decoder Network for Real-time Natural Scene Text Processing }

\author{
  Zichuan Liu$^{1}$ \texttt{zliu016@e.ntu.edu.sg} \\
  \And
  Yixing Li$^{2}$  \texttt{yixingli@asu.edu}\\
  \AND
  Fengbo Ren$^2$  \texttt{renfengbo@asu.edu}\\
  \And
  Hao Yu$^1$ \texttt{haoyu@ntu.edu.sg}\\
  \AND
  \texttt{$^1$Nanyang Technological University, $^2$Arizona State University} 
}

\begin{document}

\maketitle
\input{abstract.tex}
\input{intro.tex}
\input{section2.tex}
\input{section3.tex}
\input{section4.tex}
\input{conclusion.tex}

\newpage
	\bibliographystyle{plainnat}
	\bibliography{reference}

\end{document}

%% file: abstract.tex
\pdfoutput=1
\begin{abstract} \small
In this paper, we develop a binary convolutional encoder-decoder network (B-CEDNet) for	natural scene text processing (NSTP). It converts a text image to a class-distinguished salience map that reveals the categorical, spatial and morphological information of characters. The existing solutions are either memory consuming or
run-time consuming that cannot be applied to real-time applications on resource-constrained devices such as advanced driver assistance systems. The developed network can process multiple regions containing characters by one-off forward operation, and is trained to have binary weights and binary feature maps, which lead to both remarkable inference run-time speedup and memory usage reduction. By training with over 200, 000 synthesis scene text images (size of $32\times128$), it can achieve $90\%$ and $91\%$ pixel-wise accuracy on ICDAR-03 and ICDAR-13 datasets. It only consumes $4.59\ ms$ inference run-time realized on GPU with a small network size of 2.14 MB, which is up to $8\times$ faster and $96\%$ smaller than it full-precision version. 
\end{abstract}

%% file: intro.tex
\section{Introduction} 

The success of convolutional neuron network (CNN) has resulted in a potential general machine learning engine for various computer vision applications \cite{lecun1998gradient, krizhevsky2012imagenet} as well as for the natural scene text processing (NSTP), such as text detection, recognition and interpretation from images. Applications, such as Advanced Driver Assistance System (ADAS) for road signs with text, however require a real-time processing capability that is beyond the existing approaches \cite{jaderberg2014synthetic} from processing functionality, efficiency and latency.



A general NSTP system is based on a two-phase end-to-end pipeline that firstly segments the text region from the original image and then performs recognition of the cropped image. In recognition phase, the cropped image is transformed into text sequence that will be further processed by the Natural Language Processing (NLP) module \cite{chowdhury2003natural}. There are two categories of methods that can be applied, character-level method \cite{wang2012end, bissacco2013photoocr} and word-level method \cite{jaderberg2014synthetic, jaderberg2014deep}. The character-level method performs an individual character detection and recognition. It relies on a multi-scale sliding window strategy combined with a convolutional neural network to localize characters, resulting in long processing latency in detection \cite{wang2012end}. On the other hand, the word-level method \cite{jaderberg2014synthetic} has shorter inference time but large memory consumption due to the huge size of the fully-connected layer.\par

For a real-time NSTP application targeted for ADAS, one needs a method with memory efficiency, fast processing time as well as low power. Recent advancement in binary-constrained deep learning \cite{courbariaux2016binarynet} has opened up new opportunities for highly efficient CNN realization for the real-time NSTP. In this paper, we propose a binary convolutional encoder-decoder network (B-CEDNet) that can perform a real-time one-shot text interpretation. Instead of detecting characters sequentially \cite{bissacco2013photoocr, wang2012end, shi2015end}, our proposed network detects multiple characters simultaneously. It can distinguish different classes of characters and recovers their spatial and morphologic information by one forward pass. The output is a set of salience maps with the same size as the input image, indicating pixel-wise posterior probability distribution over a category space that composes 26-character and a background class, which allows paralleled character prediction with significant speedup. More importantly, different from traditional CNN engine, by applying binary constraints in training, it results in a B-CEDNet with massive computing parallelism with binarized weights and activations. The experiment shows that our proposed network achieves up to $91\%$ pixel-wise accuracy on public dataset \cite{lucas2003icdar, karatzas2013icdar}. Furthermore, we observe an impressive $4.59\ ms$ average forward time for processing a $32\times128$ grayscale scene text image with a small network size of $2.14$ MB, which has up to $8\times$ faster run-time and $96\%$ smaller memory usage than its full-precision version.\par

%% file: section2.tex
\pdfoutput=1
\section{B-CEDNet architecture} 

\begin{figure}[h]
	\centering
	\includegraphics[width=0.9\linewidth]{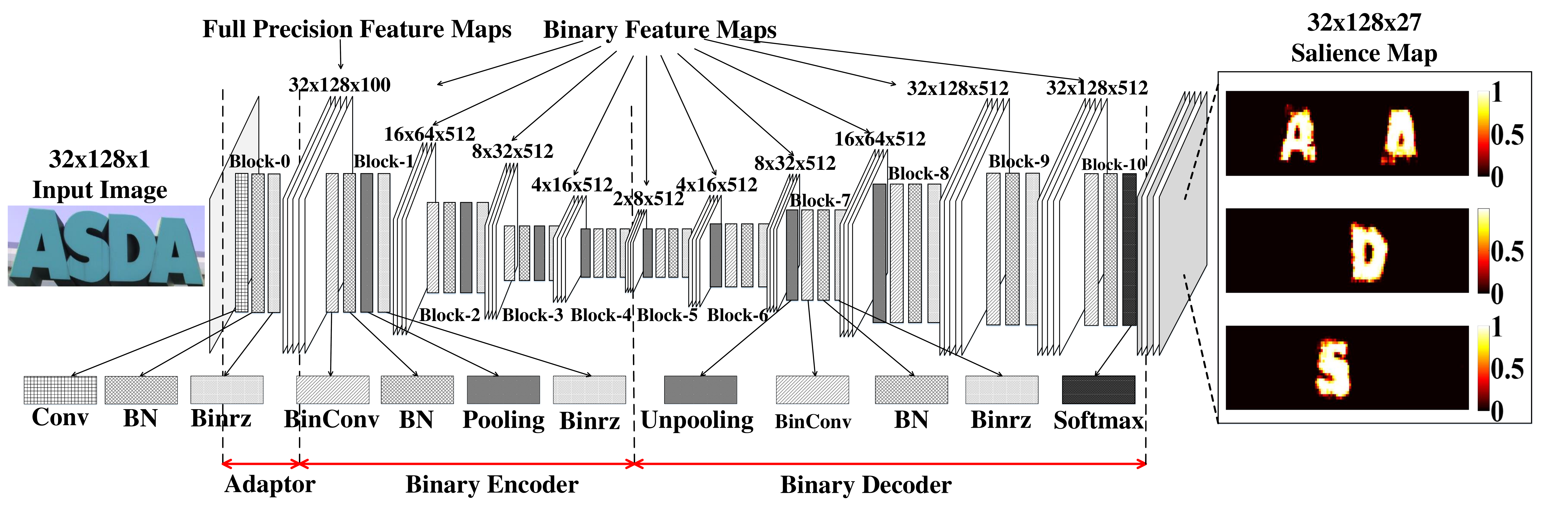}
	\caption{(a) Conventional architecture of NSTP system; (b) Proposed B-CEDNet NSTP system.}
	\label{fig:arch}
\end{figure}

The conventional NSTP architecture \cite{bissacco2013photoocr} ignores the spatial information of features, which leads to a slow sequential character detection. In contrast, the proposed B-CEDNet shown in Figure \ref{fig:arch} makes use of deconvolution techniques \cite{badrinarayanan2015segnet} to recover this spatial information, allowing paralleled character processing. Specifically, it takes a $32\times128$ grayscale image as input and converts it to a compact high-level feature, which is further decoded into a set of salience maps that indicate the categorical, spatial and morphologic information of the characters. The B-CEDNet consists of three main modules, adapter module, binary encoder module and binary decoder module. The adapter module (block-0) contains a full-precision convolutional layer, followed by a batch-normalization (BN) layer and binarization (Binrz) layer. It transforms the input data into binary format before feeding the data into binary encoder module. The binary encoder module consists of 4 blocks (block-1 to -4), each of which has one binary convolutional (BinConv) layer, one batch-normalization (BN) layer, one pooling layer and one binarization (Binrz) layer. The BinConv layer takes binary feature maps $a_{k-1}^b \in \{0,1\}^{W_{k-1}\times H_{k-1}\times D_{k-1}}$ as input and performs binary convolution operation which is illustrated as follows:
\begin{equation}
	s_k(x,y,z) = \sum_{i=1}^{w_k}{\sum_{j=1}^{h_k}{\sum_{l=1}^{d_k}{XNOR(w_k^b(i,j,l,z), a_{k-1}^b(i+x-1, j+y-1, l))}}},
\end{equation}
where $XNOR(\cdot)$ is defined as bit-wise XNOR operation, $w_k^b\in\{0,1\}^{w_k\times h_k\times d_k}$ are the binary weights in $k$-th block and $s_k \in \mathbb{R}^{W_k \times H_k \times D_k}$ is the output of the spatial convolution. Then $s_k$ is normalized by the BN layer before pooling and binarization. The output of $k$-th BN layer $a_k\in\mathbb{R}^{W_k\times H_k \times D_k}$ is represented by
\begin{equation}
	a_k(x, y, z) = \frac{s_k(x, y, z)-\mu(x,y,z)}{\sqrt{\sigma^2(x,y,z)+\epsilon}}\gamma(x,y,z)+\beta(x,y,z),
\end{equation}
where $\mu$ and $\sigma^2$ are the expectation and variance over the mini-batch, while $\gamma$ and $\beta$ are learnable parameters\cite{ioffe2015batch}. The output of the BN layer is subsequently down-sampled by the pooling layer. Here we apply $2\times2$ max-pooling to filter out the strongest response which will be binarized by the Binrz layer. The binarized activations $a_k^b$ of $k$-th block can be represented as
\begin{equation}
	a_k^b(x,y,z) = \begin{cases}
						0,\ \ \ \ a_k(x,y,z) \leq 0\\
						1,\ \ \ \ a_k(x,y,z) > 0
					\end{cases}
\end{equation} 
What is more for the decoder module, it translates the compact high-level representation $a_5^b\in\{0,1\}^{2\times 8\times 512}$ generated by the encoder into a set of salience maps $p\in \mathbb{R}^{32\times 128\times27}$ that indicates the spatial probability distribution over category space including 26 characters and a background class. The decoder module is composed of 6 convolutional blocks (block-5 to -10). Block-5 to -8 are formed by one un-pooling layer, one BinConv layer, one BN layer and one Binrz layer. 
Note that there exists a symmetric structure along block-1 to -8. Thus the un-pooling layers \cite{badrinarayanan2015segnet} within block-5 to -8 simply assign the input pixels back to their original position according to the index generated by the corresponding max-pooling layer and pad the remains with zeros. The up-sampled feature maps then go through the binary convolution, normalization and binarization. The output of block-8 is a $32\times 128\times 512$ tensor which will be processed by block-9 and -10 to generate spatial salience map. Block-9 and -10 form a 2-D spatial classifier with $1\times1$ convolution window and softmax output. It produces the posterior probability distribution over the category space for each pixel in the original image.\par

%% file: section3.tex
\pdfoutput=1
\section{Training} 

The B-CEDNet can be trained and optimized under binary constraints proposed in \cite{courbariaux2016binarynet}, which can significantly reduce memory usage and also improve parallelism. In the existing binary CNN method \cite{courbariaux2016binarynet}, hinge loss function is used for training, which is however unsuitable for our application because it fails to provide a probability interpretation from the input image. In this paper, we apply cross-entropy error as the loss function by removing the Binrz layer in block-10. For our application, the prediction error $J$ is represented as follows:
\begin{equation}
	J(w) = -\frac{1}{N\cdot W_{10}\cdot H_{10} }\sum_{i=1}^N\sum_{m=1}^{W_{10}}\sum_{n=1}^{H_{10}}{\sum_{c=1}^{C}{[\textbf{1}\{Y^{(i)}(x,y)=c\}\ln\frac{e^{a_{10}(m,n,c)}}{\sum_{l=1}^{C} {e^{a_{10}(m,n,l)}}}]}},
\end{equation}
where $N$ is the number of training sample in a mini-batch, $C$ is the number of classes, $w$ is the filter weights, $Y^{(i)}\in\{1,..., C\}^{H_{10}\times W_{10}}$ is the 2-D label of $i$-th training image, $a_{10}\in\mathbb{R}^{H_{10}\times W_{10}\times C}$ \footnotemark[1] is the output of the BN layer in block-10.\par

To achieve generality of trained model, it usually needs a large amount of labeled data for training. However, the existing datasets are limited to word-level annotation \cite{veit2016cocotext} or cannot provide enough pixel-wise labeled data \cite{karatzas2013icdar}. Inspired by \cite{jaderberg2014synthetic}, we create a text rendering engine that generates texts with different fonts, colors and projective distortions. The labeled image has the same size with the corresponding text image and provides a pixel-wise labeling over the category space. 
Additionally, our model is trained by AdaMax optimizer \cite{kingma2014adam} with initial learning rate of $0.002$, learning rate decay of $0.9$ and mini-batch size of $20$.

\footnotetext[1]{In the proposed network $C$, $H_{10}$, $W_{10}$ are $27$, $32$ and $128$, respectively.}

%% file: section4.tex
\pdfoutput=1
\section{Experiment results and discussion} 
\subsection{Experiment setup}
The model is built based on MatConvNet\cite{vedaldi15matconvnet} on Dell Precision T7500 server with Intel Xeon 5600 processor, 64GB memory and TITAN X GPU. 
To evaluate the performance, we train the model on synthesis scene text dataset with 200, 000 images and test it on the standard datasets ICDAR-03 and ICDAR-13 \cite{lucas2003icdar, karatzas2013icdar}.

\subsection{Pixel-wise accuracy}

\begin{figure*}[h]
	\centering
	\includegraphics[width=0.8\linewidth]{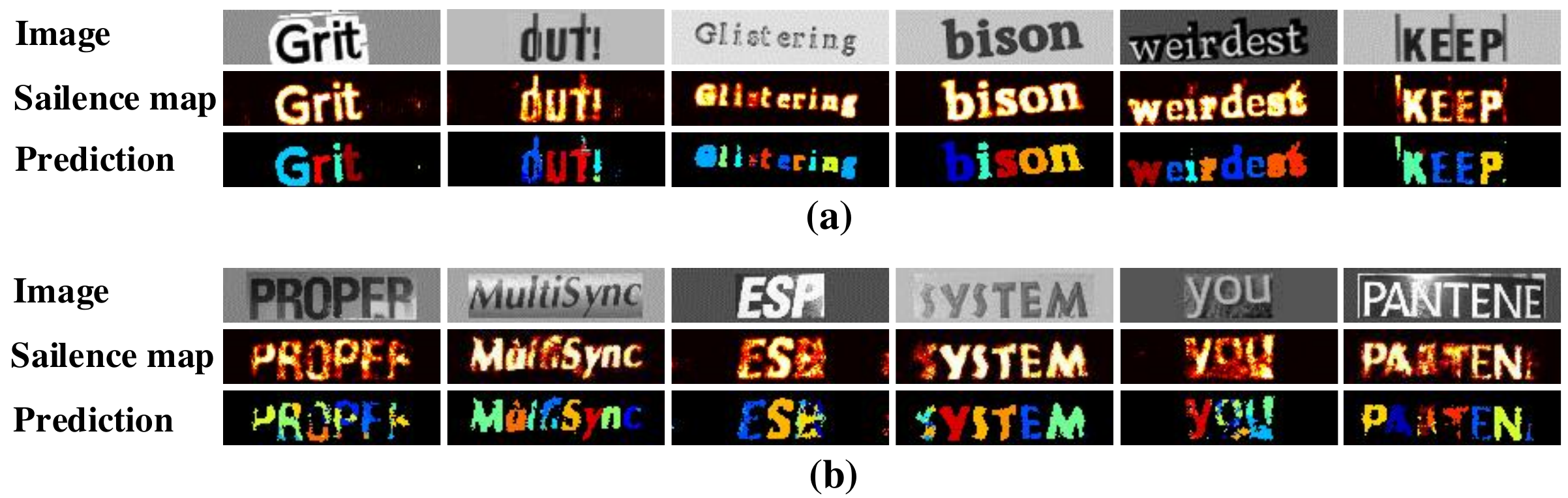}
	\caption{(a) Testing samples with high prediction accuracy rate; (b) Testing samples with low prediction accuracy rate.}
	\label{fig:test_sample}
\end{figure*}

Table \ref{tab:acc} reports the pixel-wise accuracy on the testing datasets mentioned above. We achieve notable $90\%$ and $91\%$ pixel-accuracy on ICDAR-03 and ICDAR-13. 
Figure \ref{fig:test_sample} (a) shows the ideal cases, where our binary model is capable to precisely recognize each character and recover their outline information under normal contrast and even illumination condition. While, in the non-ideal cases, the salience maps produce low confidence value in the area with uneven illumination or low contrast as shown in Figure \ref{fig:test_sample} (b). It indicates that the model feels confused (unconfident) when interpreting the characters.
It is worth noting that the robustness can be improved by extending the training set to include the missed cases. Therefore we believe B-CEDNet will be promising in fast scene text detection and natural text interpretation.

\input{accuracy.tex}

\subsection{Run-time and memory usage}

\begin{figure*}[h]
	\centering
	\includegraphics[width=0.8\linewidth]{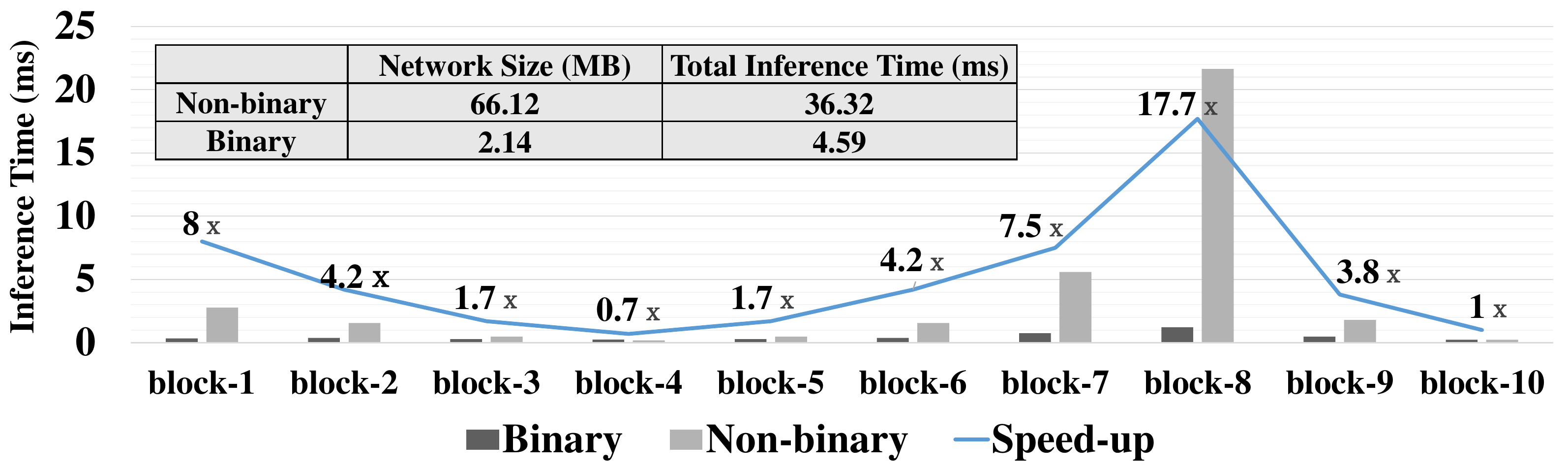}
	\caption{Inference run-time and memory consumption of B-CEDNet and its full-precision version.}
	\label{fig:speed}
\end{figure*}

Figure \ref{fig:speed} compares the inference time for B-CEDNet running on baseline kernel and XNOR kernel \cite{courbariaux2016binarynet}. Baseline kernel is an optimized matrix multiplication kernel, while the XNOR kernel is tailored for bit-count operation in binary network. We measure the inference time with a batch of input images (size of 16) to obtain higher utilization of GPU. Due to the bit-count operation and huge memory access reduction, the B-CEDNet achieves an average of 4.59 ms inference time and $8\times$ speedup with XNOR kernel on TITAN X GPU compared with baseline kernel. Since XNOR kernel introduces run-time overhead concatenating 32 binary values into a 32-bit register, the speedup is more remarkable when the convolutional operation becomes more intensive. Accordingly, the most computation-intensive block, block-8 has the highest speedup of $17.7\times$. On the other hand, the memory usage is reduced by over $96\%$ ($2.14$ MB) compared with the full-precision version ($66.12$ MB). 


%% file: accuracy.tex
\pdfoutput=1
\begin{table}[h]\renewcommand{\arraystretch}{1.2}
	\centering
	\caption{Pixel-wise accuracy our proposed network on ICDAR-03 and ICDAR-13.}
	\label{tab:acc}
	\scalebox{0.8}{
	\begin{tabular}{|c|cc|}
		\hline
		& \textbf{ICDAR-03} & \textbf{ICDAR-13} \\ \hline
		\textbf{Binary} & \textbf{0.90}     & \textbf{0.91}          \\ \hline
		\textbf{Non-binary} & \textbf{0.90}     & \textbf{0.91}          \\ \hline
	\end{tabular} }
\end{table}

%% file: conclusion.tex
\pdfoutput=1
\section{Conclusion}

In this paper, we have developed a binary convolutional encoder-decoder network (B-CEDNet) for real-time NSTP applications. The B-CEDNet can effectively capture the categorical, spatial and morphologic information from the text image and is extremely computationally efficient with binary weights and activations trained from binary constraints. By training with over 200, 000 synthesis scene text image, it can achieve up to $91\%$ pixel-wise accuracy. It only consumes $4.59\ ms$  inference time realized on GPU with a small network size of $2.14$ MB, which is up to $8\times$ faster and $96\%$ smaller than its non-binary version.